# Navigating Heat Exposure: Simulation of Route Planning Based on Visual Language Model Agents

Haoran Ma[1], Kaihan Zhang[2] and Jiannan Cai[3(✉)]

**Abstract.** Heat exposure significantly influences pedestrian routing behaviors. Existing methods such as agent-based modeling (ABM) and empirical measurements fail to account for individual physiological variations and environmental perception mechanisms under thermal stress. This results in a lack of human-centred, heat-adaptive routing suggestions. To address these limitations, we propose a novel Vision Language Model (VLM)-driven Persona-Perception-Planning-Memory (PPPM) framework that integrating street view imagery and urban network topology to simulate heat-adaptive pedestrian routing. Through structured prompt engineering on Gemini-2.0 model, eight distinct heat-sensitive personas were created to model mobility behaviors during heat exposure, with empirical validation through questionnaire survey. Results demonstrate that simulation outputs effectively capture inter-persona variations, achieving high significant congruence with observed route preferences and highlighting differences in the factors driving agents' decisions. Our framework is highly cost-effective, with simulations costing \$0.006 and taking 47.81s per route. This Artificial Intelligence-Generated Content (AIGC) methodology advances urban climate adaptation research by enabling high-resolution simulation of thermal-responsive mobility patterns, providing actionable insights for climate-resilient urban planning.

[1] Haoran Ma
Institute of Space and Earth Information Science, The Chinese University of Hong Kong, Hong Kong SAR, China
e-mail: haoranma@cuhk.edu.hk

[2] Kaihan Zhang
Cho Chun Shik Graduate School of Mobility, Korea Advanced Institute of Science & Technology, South Korea
e-mail: kaihn@kaist.ac.kr

[3] Jiannan Cai (✉)
College of Survey and Geo-informatics, Tongji University, China
e-mail: jncai@tongji.edu.cn

# 1 Introduction

Under the global climate change crisis, heat exposure have emerged as a critical public health challenge (Ebi et al., 2021). Data from the United Nations Environment Programme (UNEP) reveals that heat exposure contributes to over 100,000 excess deaths annually, with this number escalating alongside rising greenhouse gas emissions (Masson-Delmotte et al., 2021). The 1.5°C temperature control target outlined in the Paris Agreement faces unprecedented pressure, necessitating urgent refinement in urban microclimate governance. In this context, analyzing the mobility behaviors of diverse social groups in heat environments and establishing scientific heat exposure avoidance strategies have become critical for enhancing urban climate resilience and public well-being.

Traditional research on heat exposure mobility behaviors predominantly relies on resource-intensive methodologies. For instance, subjective thermal perception data collected through questionnaires suffer from recall bias and sample limitations (He et al., 2023). While Ecological Momentary Assessment (EMA) captures real-time psychological feedback, it requires continuous device deployment and participant compliance (Stone et al., 2023). Similarly, GPS trajectory tracking can map movement patterns but raises privacy concerns (Yang et al., 2025b). These approaches face significant bottlenecks in terms of data collection efficiency, scalability, and ethical compliance. These limitations lead to an inability to precisely capture path choices among different individuals, thus reducing the effectiveness of huma-centred, heat-adaptive routing suggestions.

Recent advances in Artificial Intelligence Generated Content (AIGC) technologies have opened new avenues for human behavioral simulation. For example, the Stanford Town project has explored social interactions and daily routines simulated by Large Language Model (LLM) agents (Park et al., 2023). In addition, the enhanced visual capabilities of LLM (i.e., Vision Language Model, VLM) now enable precise analysis of urban attractiveness in urban environment, achieving a high correlation with human perception (Malekzadeh et al., 2025). In addition, Street View Imagery (SVI), a human-centric, fine-grained built-environment dataset, has been widely used in urban thermal environment studies (Yang et al., 2025a). For instance, Hu et al. (2024) utilized 3D morphological data from SVI for street-level thermal mapping. These advances of method and dataset provide unprecedented opportunities to investigate mobility behaviors of different social groups under heat exposure.

In this study, we propose an innovative VLM-driven agent framework to simulate pedestrian routing behaviors across different populations under heat exposure through vision-language fusion. This methodology offers dual benefits: exploiting the visual cognition of VLMs to dynamically assess thermal risks along routes, while using the social attribute modeling capabilities of VLMs to generate differentiated behavioral strategies without compromising privacy. Our method delivers a scalable digital solution for climate-resilient urban planning.

## 2 Methodology

Our VLM-driven Persona-Perception-Planning-Memory (PPPM) framework establishes a closed-loop system in which four functionally distinct yet interconnected modules collaboratively simulate pedestrian routing behaviors (Fig. 1). The Persona module generates population profiles by embedding social attributes into the VLM to create differentiated agent prototypes that directly condition the Perception module to evaluate thermal comfort levels use SVI. Further, the Planning module dynamically weights the perceptual outputs with the shortest path calculated by the Dijkstra algorithm to obtain the Top-K candidate paths. Finally, the Memory module optimizes the agent's decision-making strategies through an experience replay mechanism.

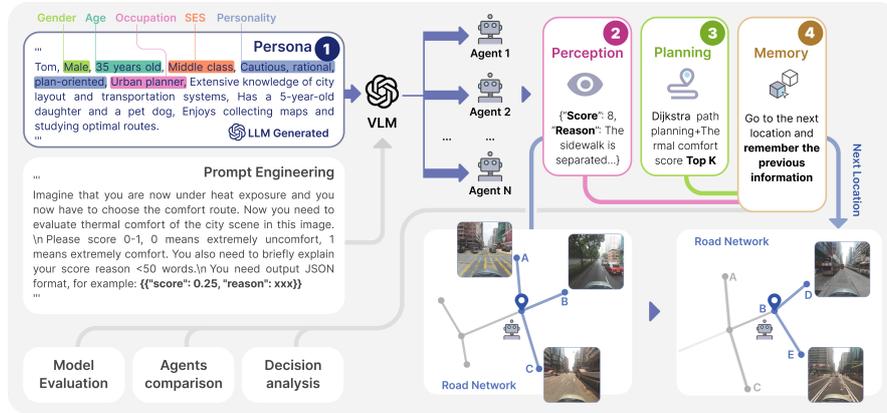

**Fig. 1.** Persona-Perception-Planning-Memory (PPPM) framework.

### 2.1 Explanation of Each Module

**Persona module.** This module creates population profiles via ChatGPT, incorporating attributes such as gender, age, income, and personality traits. Prior studies have demonstrated significant correlations between these demographic variables and mobility behaviors under heat exposure (Nazarian & Lee, 2021). This study we created eight heat-sensitive personas (Table 1), which are input into the Perception module as part of prompt condition.

**Table 1.** Profiles of the eight heat-sensitive personas.

| Name | Gender | Age | Income Level | Occupation |
|---|---|---|---|---|
| Alex | Male | 31 | Middle | Graphic designer |
| Bob | Male | 68 | Low-middle | Retired worker |
| Emma | Female | 75 | High | Retired teacher |
| Lisa | Female | 42 | High | Company CEO |
| Maria | Female | 55 | Low-middle | School janitor |
| Ryan | Male | 23 | Middle | Software engineer |
| Sara | Female | 28 | Low | Hospital nurse |
| Tom | Male | 35 | Middle | Urban planner |

**Perception module.** Prompt engineering guides the VLM to play specific social roles within heat exposure scenarios (details of prompt are shown in Fig. 1). SVI, acquired through Baidu Map API, is evaluated by the fine-tuned VLM to quantify thermal comfort scores (range from 0-1). Model output parameters are configured with a temperature of 0.8 and a maximum token length of 300 to balance generative stability and diversity, respectively. Through experimentation, Gemini-2.0 has the highest accuracy and the lowest response time and is selected as the baseline model. (Table 2).

**Table 2.** Model comparation analysis.

| Model | Avg. Accuracy* | Avg. Cost ($) | Avg. Time (s) |
|---|---|---|---|
| ChatGPT-4o | 70.0% | 0.049 | 84.82 |
| Claude-3.5 | 80.0% | 0.045 | 51.46 |
| Gemini-2.0 | **100.0%** | **0.006** | **47.81** |

* The accuracy (%) of reaching the home location accurately in 10 simulations.

**Planning module.** In this study, urban road networks are represented as graph structures where nodes store spatial coordinates, SVI retrieval interfaces, and thermal metadata. A dynamic planning algorithm integrates Dijkstra's shortest-path weights with thermal comfort scores, producing Top-K candidate routes that optimize both travel efficiency and heat avoidance.

**Memory module.** This module employs a trajectory summary mechanism to iteratively refine agent decision-making strategies. By systematically recording historical path parameters (e.g., environmental selection biases), the module constructs a behavioral pattern knowledge base. This dual-function system not only enhances agent decision ability but also provides a structured dataset for analyzing behavioral patterns of heat-sensitive populations.

### 2.2 Model Evaluation and Explanation

Agent-based stimulation results often face doubts about authenticity and explainability. To address first issue, we introduce a dual-dimension evaluation metrics to verify the authenticity of our framework. We firstly collect real-world path selection data via questionnaires. The Path Overlap Index (POI) quantifies the similarity between simulation results and real choices, considering path length and turning frequency. Second, respondents rate thermal comfort in 42 typical scenarios (1-5 scale). We compare these ratings with those from VLM agents using Pearson correlation analysis as the Perception Consistent Index (PCI). 20 participants were recruited, with socio-demographic attribute proportions same to those in Table 1.

In addition, the explainability of agents decision-making logic is important. During each path choice, the system requires agents to generate a decision rationale text (less than 50 words), accumulating 896 explanation records. We use a two-stage topic mining method. First, DeepSeek-R1 model performs semantic vectorization clustering to identify common decision factors. Then, we extract cross-group differential keywords according to their frequency, forming a heat-sensitive behavioral decision-making atlas.

## 3 Results

This study was conducted in Hong Kong. There, the summer ground temperature (average 30°C) in the core area is 5.2°C higher on average than in the suburbs (Lou et al., 2024), creating a typical scenario for our study (Fig 2). We particularly focused on heat exposure during the last mile commute from metro stations to homes, which have been a critical concern in heat exposure research.

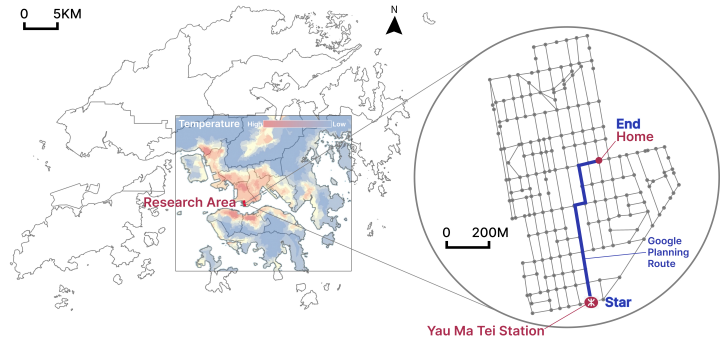

**Fig. 2.** Study areas.

### 3.1 Heterogeneous Agent-Specific Routing Patterns

Our analysis revealed significant heterogeneity in pedestrian routing behaviors across different agents. As illustrated in Fig. 3, under identical origin and destination conditions, eight agents exhibited distinct path length variations. We classified mobility patterns into three typologies: 1) **Conservative routing** (Tom, Sara, Maria): These agents preferred stable trajectories with average distances of 1,050-1,060 meters, maintaining detour ranges of 150-160 meters. 2) **Exploratory rout-** **ing** (Lisa, Emma): Characterized by maximal path deviations exceeding 1,260 meters, potentially reflecting proactive shade-seeking exploration. 3) **Balanced routing** (Bob, Alex): Achieved intermediate performance, optimizing the trade-off between route efficiency (1,100-1,150 meters) and heat exposure minimization.

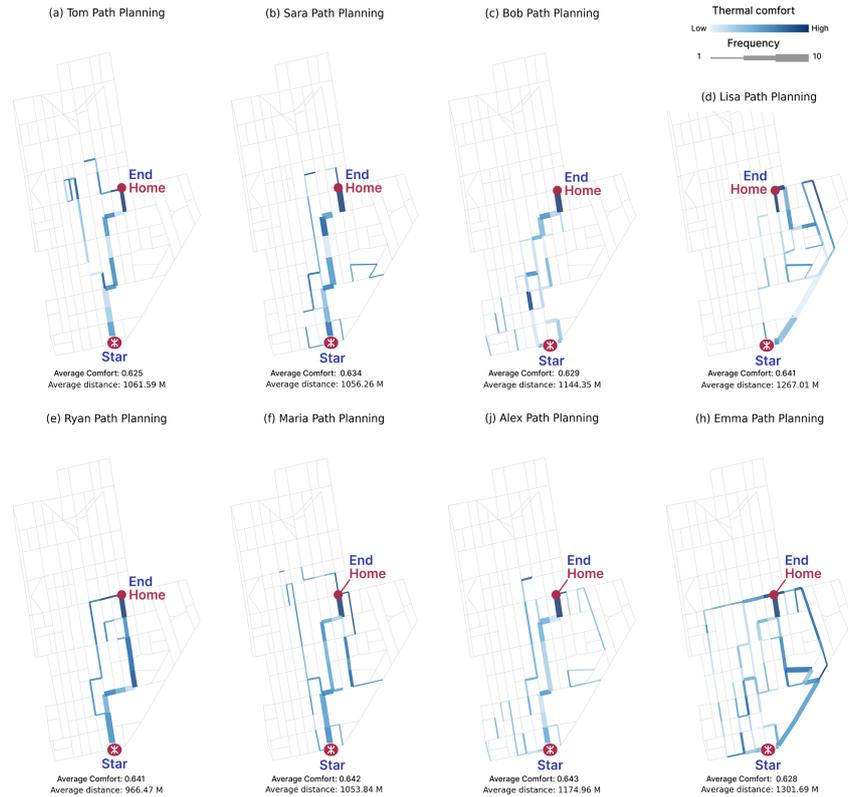

**Fig. 3.** Comparative analysis of pedestrian routing patterns for different agents.

## 3.2 Demographic Disparities in Routing Behaviors

We compared the pedestrian routing patterns of different demographic groups under heat exposure scenarios (Fig. 4). Specifically, from a gender perspective, the average travel distance for females (1081.68 meters) was slightly longer than that for males (971.11 meters), with their routes exhibiting more twists, turns, and circuitous features. This may reflect females' tendency to prioritize shaded pathways during heat exposure, even if such choices increase travel distance. Additionally, from an age-based analysis, travel distance showed a significant upward trend with increasing age. The average distance for young groups was 1096.75 meters, while senior groups reached 1273.50 meters. This disparity may stem from older populations requiring more detours to avoid heat risks.

In addition, high-income groups exhibited the longest travel distances (1267.74 meters), while low-income groups had shortest (1056.26 meters). This phenomenon may indicate that high-income individuals enjoy freedom in route selection, whereas low-income groups prioritize efficiency through direct paths. Low-and lower-middle-income groups demonstrated moderate travel distances.

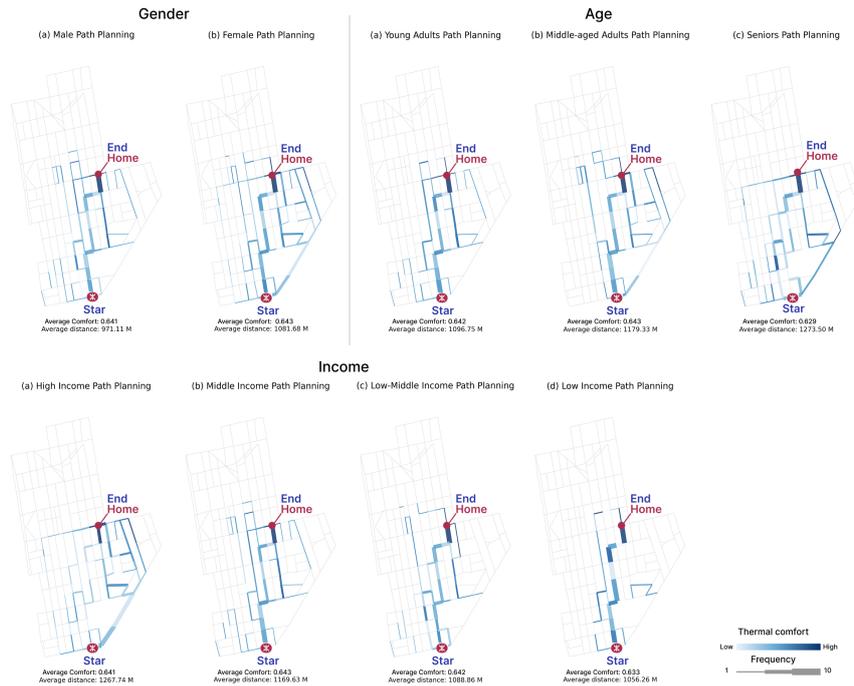

**Fig. 4.** Comparative analysis of pedestrian routing patterns for different groups.

## 3.3 Thermal Perception Dynamics and Model Validation

We conducted an in-depth analysis of thermal comfort perception across different groups under identical route conditions (Fig. 2). At the individual level (Fig. 5a), significant differences in thermal comfort scores were observed among virtual agents. Ryan exhibited the highest score, while Bob scored relatively lower. Notably, most agents' scores fluctuated around the average (0.60), indicating strong model stability and also capturing individuals heterogeneously.

From a group characteristic perspective, gender differences were minimal (Fig. 5b), with no significant disparity in thermal comfort scores between males and females. Age-based analysis (Fig. 5c) revealed a clear declining trend: younger groups demonstrated the highest thermal comfort scores, while older groups scored significantly lower. Income-level analysis (Fig. 5d) showed that middle-income groups achieved the highest thermal comfort scores, whereas high- and low-income groups scored lower. This distribution contrasts intriguingly with previously observed mobility patterns, suggesting that travel distance does not directly determine perceived thermal comfort.

To validate model realism, we collected real-world route selection data through questionnaires and quantified simulation-to-reality similarity using POI and PCI metrics. Results demonstrated high consistency overall, though simulations for older adults and high-income groups showed poorer alignment, warranting focused analysis in future studies (Fig. 5e).

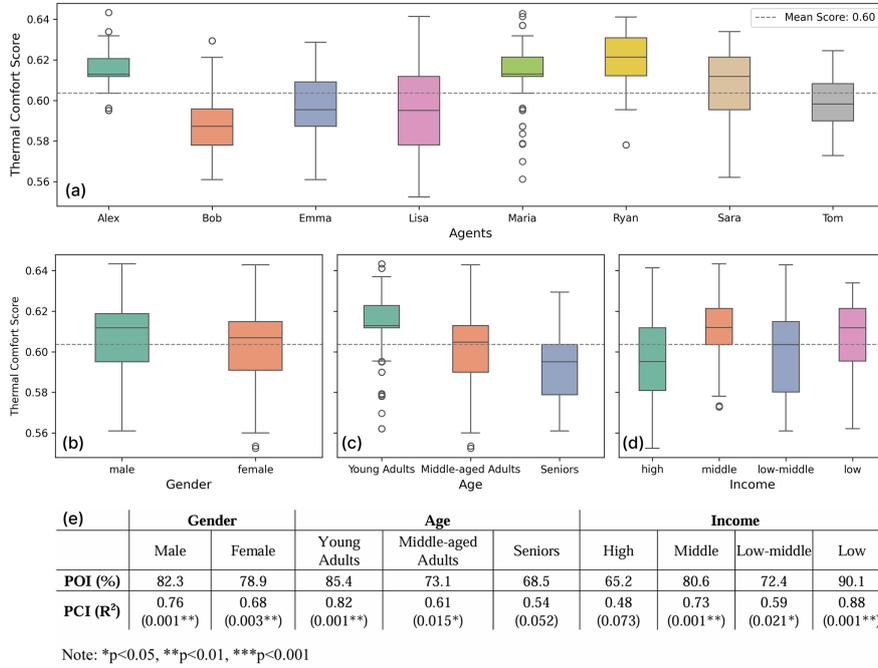

| (e) | Gender | | Age | | | Income | | | |
|---|---|---|---|---|---|---|---|---|---|
|  | Male | Female | Young Adults | Middle-aged Adults | Seniors | High | Middle | Low-middle | Low |
| **POI (%)** | 82.3 | 78.9 | 85.4 | 73.1 | 68.5 | 65.2 | 80.6 | 72.4 | 90.1 |
| **PCI ($R^2$)** | 0.76 (0.001**) | 0.68 (0.003**) | 0.82 (0.001**) | 0.61 (0.015*) | 0.54 (0.052) | 0.48 (0.073) | 0.73 (0.001**) | 0.59 (0.021*) | 0.88 (0.001**) |

Note: *$p<0.05$, **$p<0.01$, ***$p<0.001$

**Fig. 5.** Comparison of thermal perception results and validation of model.

### 3.4 Environmental Priority in Decision Making

To deeply understand decision-making mechanisms across different agents in route planning, we deconstructed the model's decision processes using topic analysis. The study identified seven key environmental feature dimensions: urban structure, microclimate conditions, sun exposure & shading, surface materials, traffic & vehicles, green infrastructure, and comfort perception (Fig. 6).

Analysis revealed that sun exposure & shading dominated decision weights across all agents (24.2%-35.9%), indicating that avoiding direct sunlight is the primary consideration under extreme heat conditions. Notably, Bob exhibited the highest emphasis on this factor (35.9%), while Emma showed the lowest (24.2%). Urban structure emerged as the second most influential factor, with relatively uniform weight distribution among agents (10.2%-21.8%). Maria demonstrated the strongest focus on urban structure (21.8%), which may reflect active prioritization of architectural shelter effects.

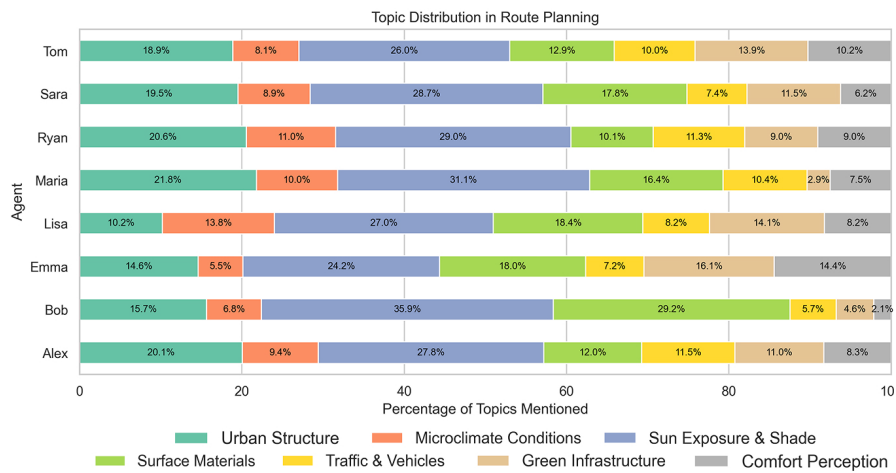

**Fig. 6.** Distribution of topics mentioned by each agent.

## 4 Conclusion and Discussion

This study pioneers the integration of VLMs into heat exposure research through the proposed PPPM framework. By synergizing AIGC capabilities with human-centric SVI, our approach overcomes traditional limitations in data collection efficiency and ethical compliance while enabling fine-grained simulation of heterogeneous population pedestrian routing behaviours. Three major advancements emerge: 1) Our framework captures individual and demographic-specific heterogeneity, identifying three major pedestrian routing patterns; 2) High consistency between simulated and empirical data confirms our framework's authenticity; and 3) Through explanation analysis, Sun exposure & shading dominates route choices, providing empirical validation for climate-responsive urban design principles. In summary, our research offers a digital solution for exploring human-centred, heat-adaptive routing suggestions. It is valuable for enhancing urban climate resilience and human well-being.

However, there some limitations. The current SVI-based perception module focuses on static thermal features, neglecting dynamic meteorological factors like wind speed and humidity. Future iterations could integrate real-time weather APIs for enhanced environmental fidelity. Additionally, our study did not consider more transportation modes and time factors (such as the evening rush hour). Future research needs to be further refined. Moreover, the evaluation scope concentrates on Hong Kong's high-density urban form. Cross-validation across diverse city typologies is needed to assess generalizability.

# 5 References

EBI, K. L., CAPON, A., BERRY, P., BRODERICK, C., DE DEAR, R., HAVENITH, G., HONDA, Y., KOVATS, R. S., MA, W., MALIK, A., MORRIS, N. B., NYBO, L., SENEVIRATNE, S. I., VANOS, J. & JAY, O. 2021. Hot weather and heat extremes: health risks. *Lancet,* 398**,** 698-708.

HE, B. J., WANG, W., SHARIFI, A. & LIU, X. 2023. Progress, knowledge gap and future directions of urban heat mitigation and adaptation research through a bibliometric review of history and evolution. *Energy and Buildings,* 287.

HU, Y., QIAN, F., YAN, H., MIDDEL, A., WU, R., ZHU, M., HAN, Q., ZHAO, K., WANG, H., SHAO, F. & BAO, Z. 2024. Which street is hotter? Street morphology may hold clues - thermal environment mapping based on street view imagery. *Building and Environment,* 262**,** 111838.

LOU, S. W., FENG, C., ZHANG, D. Q., ZOU, Y. K. & HUANG, Y. 2024. Heat exposure inequalities in Hong Kong from 1981 to 2021. *Urban Climate,* 56.

MALEKZADEH, M., WILLBERG, E., TORKKO, J. & TOIVONEN, T. 2025. Urban attractiveness according to ChatGPT: Contrasting AI and human insights. *Computers Environment and Urban Systems,* 117.

MASSON-DELMOTTE, V., ZHAI, P., PIRANI, A., CONNORS, S. L., PéAN, C., CHEN, Y., GOLDFARB, L. & GOMIS, M. I. 2021. Climate Change 2021—The Physical Science Basis. *Chemistry international,* 43**,** 22-23.

NAZARIAN, N. & LEE, J. K. W. 2021. Personal assessment of urban heat exposure: a systematic review. *Environmental Research Letters,* 16.

PARK, J. S., O'BRIEN, J., CAI, C. J., MORRIS, M. R., LIANG, P. & BERNSTEIN, M. S. 2023. Generative Agents: Interactive Simulacra of Human Behavior. *UIST '23: Proceedings of the 36th Annual ACM Symposium on User Interface Software and Technology***,** 2 (22 pp.)-2 (22 pp.).

STONE, A. A., SCHNEIDER, S. & SMYTH, J. M. 2023. Evaluation of Pressing Issues in Ecological Momentary Assessment. *Annual Review of Clinical Psychology,* 19**,** 107-131.

YANG, S., CHONG, A., LIU, P. & BILJECKI, F. 2025a. Thermal comfort in sight: Thermal affordance and its visual assessment for sustainable streetscape design. *Building and Environment,* 271**,** 112569.

YANG, W., ZHANG, G. Y., LIU, Y. & AN, Z. H. 2025b. Heat exposure assessment and comfort path recommendations for leisure jogging based on street view imagery and GPS trajectories. *Sustainable Cities and Society,* 119.